\pdfoutput=1

\documentclass[11pt]{article}

\usepackage[final]{acl}
\usepackage{enumitem}
\usepackage{times}
\usepackage{latexsym}
\usepackage{float}
\usepackage{booktabs}
\usepackage{multirow}
\usepackage{amsmath}

\usepackage[T1]{fontenc}

\usepackage[utf8]{inputenc}

\usepackage{microtype}

\usepackage{inconsolata}

\usepackage{graphicx}

\title{Multi-Frame Vision-Language Model for Long-form Reasoning in \\Driver Behavior Analysis}

\author{
  Hiroshi Takato$^1$\thanks{CEO, Teatis inc.}, 
  Hiroshi Tsutsui$^1$,
  Komei Soda$^2$,
  Hidetaka Kamigaito$^3$ \\
  $^1$Teatis inc., $^2$Queensland university of technology, \\$^3$Nara Institute of Science and Technology (NAIST)\\
  \texttt{\{tak, james\}@dotsfty.com}, \texttt{komei.soda@connect.qut.edu.au}, \\
  \texttt{kamigaito.h@is.naist.jp}
}

\begin{document}

\maketitle

\begin{abstract}
Identifying risky driving behavior in real-world situations is essential for the safety of both drivers and pedestrians. However, integrating natural language models in this field remains relatively untapped. To address this, we created a novel multi-modal instruction tuning dataset and driver coaching inference system. Our primary use case is dashcam-based coaching for commercial drivers. The North American Dashcam Market is expected to register a CAGR of 15.4 percent from 2022 to 2027.
Our dataset enables language models to learn visual instructions across various risky driving scenarios, emphasizing detailed reasoning crucial for effective driver coaching and managerial comprehension.
Our model is trained on road-facing and driver-facing RGB camera footage, capturing the comprehensive scope of driving behavior in vehicles equipped with dashcams.
\end{abstract}

\section{Introduction}

In recent years, the dashcam monitoring industry has experienced rapid advancements, with 4 million commercial fleets now implementing dashcams and billions of data points being recorded annually in the USA \cite{dcms2024}.
This advancement facilitates the research of using Large-scale Vision Language Models (LVLMs) \cite{zhou2023driving} to interpret driving scenes, including complex city traffic situations \cite{xu2023drivegpt4,yang2023lidar,jin2023adapt,hu2023gaia1generativeworldmodel}.
In contrast to the direction, LVLMs' interpretation of both drivers' behaviors and driving actions remains unexplored, even though it is crucial not only for effective driver coaching but also for making the underlying causes of risky behaviors comprehensible to managers.

In this paper, we introduce the Multi-Frame Vision-Language Model for Reasoning in Driver Behavior Analysis, which extends the concept of visual instruction tuning \cite{liu2023visual} into the domain of risky driving behavior analysis. In analyzing the driving behavior of commercial drivers, it is essential to consider both the road-facing and driver-facing cameras. The road-facing camera captures external conditions and events, while the driver-facing camera detects distractions, aggressive reactions, and other behaviors, providing a comprehensive understanding of the driving situation. We focused on the need for road-facing RGB camera video footage and driver-facing RGB camera video footage captioning datasets. These datasets connect visual modality and language within visual instruction tuning. To address this gap, we created a high-quality video instruction dataset with complex reasoning and detailed descriptions. We also performed cross-video analysis for driver coaching.

\begin{figure}
    \centering
    \includegraphics[width=\columnwidth]{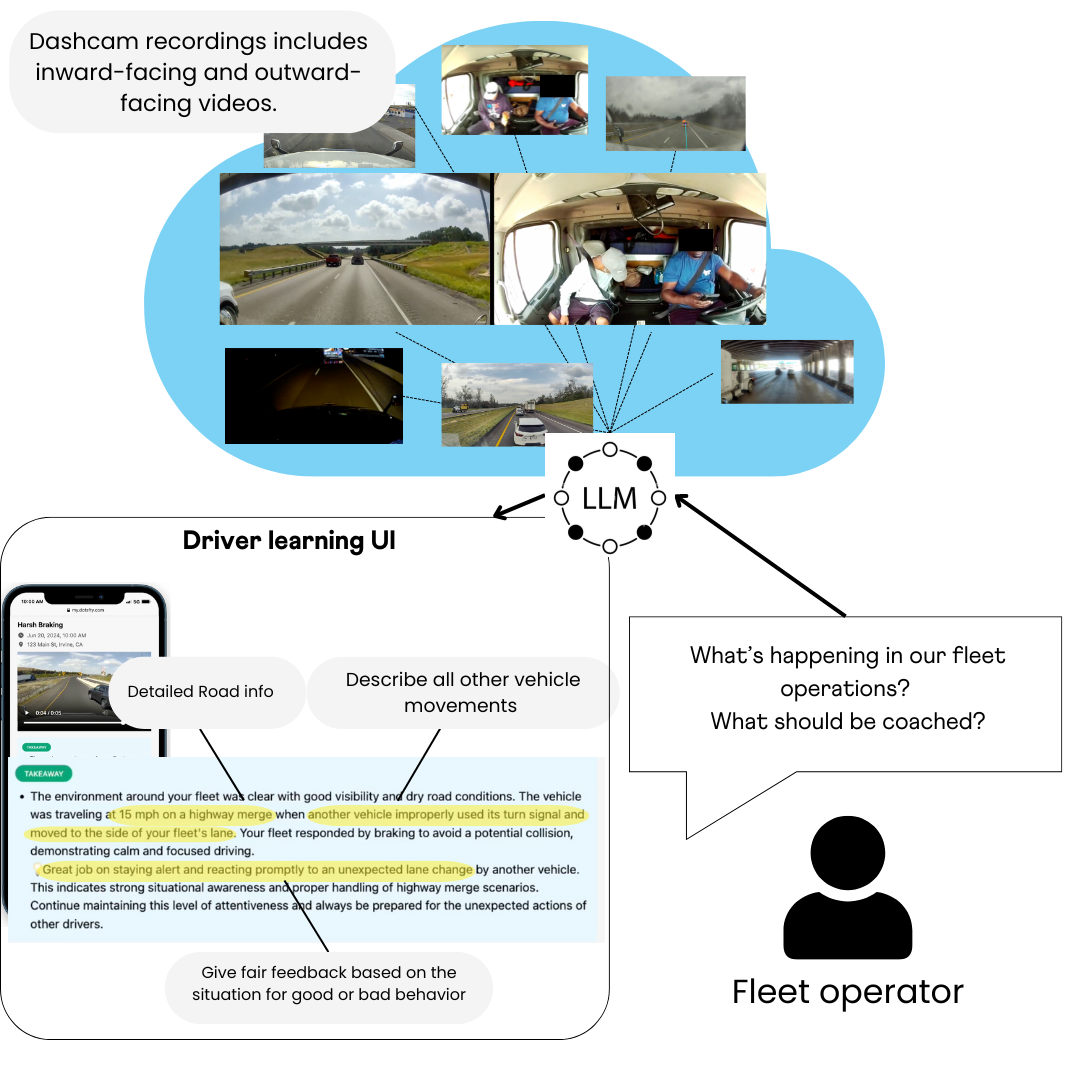}
    \caption{Overview of our targeting coaching task.}
    \label{fig:overview}
\end{figure}
\begin{table*}[t]
    \centering
    \small
    \resizebox{\textwidth}{!}{
    \begin{tabular}{p{8cm}clc}
        \toprule
        \multicolumn{4}{c}{\textbf{Event Recognition}}\\
        \midrule
        \textbf{Question} & \textbf{Answer} & \textbf{Follow-up Question (FUQ)} & \textbf{FUQ's Answer} \\
        \midrule

        Did a lane cut off happen in the video? & Yes/No & Did the car that cut into my lane use a turn signal & Yes/No \\ 

        \midrule

        \multirow{3}{*}{Can you see a driver?} & \multirow{3}{*}{Yes/No} & Is the driver smoking? & Yes/No \\
         & & Is the driver using a phone? & Yes/No \\
         & & Are there signs of aggressive reaction?  & Yes/No \\

         \midrule

         Can you see a stop sign in the video? & Yes/No & Did the ego-car ignore a stop sign? & Yes/No \\

         \midrule

         Does the ego-car maintain the safe following distance? & Yes/No & How was the speed managed for the ego-car? & Yes/No \\

         \midrule

         Did the ego-car break hard? & Yes/No & Why did the ego-car break hard? & Yes/No \\

         \midrule

         Did lane change happen in the video? & Yes/No & Why did the ego-car change a lane? & Yes/No \\

         \midrule

         Did the ego-car make a sharp turn? & Yes/No & Why does the ego-car make a sharp turn? & Yes/No \\

         \midrule

         What is the road condition \underline{Dry}, \underline{Wet} or \underline{Icy}? & \textit{Underline} & --- & --- \\

         \midrule

         What is the weather condition \underline{Clear}, \underline{Rainly}, \underline{Foggy} or \underline{Snowy}? & \textit{Underline} & --- & --- \\

         \midrule

         How is the visibility \underline{Clear}, \underline{Moderate}, Poor or Night? & \textit{Underline} & --- & --- \\

         \midrule

         What is the road information? Choose from below
\underline{Highway}, \underline{Highway Merge}, \underline{Local road}, \underline{Intersection}, \underline{3-leg intersection}, \underline{School Zone
Construction Zone}, \underline{Residential Area}, \underline{Rural Roads}, \underline{Tunnel}, \underline{Pedestrian crossroad} & \textit{Underline} & --- & --- \\

        \bottomrule
        \toprule

        \multicolumn{4}{c}{\textbf{Open Question}}\\
        \midrule

        \multicolumn{2}{l}{\textbf{Question}} & \multicolumn{2}{l}{\textbf{Answer}} \\
        \midrule

        \multicolumn{2}{l}{What is happening in the video?} & \multicolumn{2}{l}{\textit{Explanation}} \\
        \midrule

        \multicolumn{2}{l}{What driving action is recommended for the ego-car?}  & \multicolumn{2}{l}{\textit{Explanation}} \\
        
        \bottomrule
    \end{tabular}}
    \caption{Instruction templates. \textit{Underline} indicates that the answers should be chosen from the underlined texts.}
    \label{tab:instructions}
\end{table*}

In summary, our contributions can be summarized as follows:\\
\noindent(1) The development of a comprehensive visual guidance coordination dataset for both road-facing and driver-facing cameras in simultaneous recording situations.
\\
\noindent(2) Building our model, Multi-Frame Vision-Language Model by instruction-tuning Video-LLaMA \cite{zhang-etal-2023-video} on our created dataset.
\\
\noindent(3) Improving the inference ability for driver coaching in long-form detailed explanation.

\section{Task and Evaluation}

\subsection{Task Settings}

Figure \ref{fig:overview} shows our coaching task's overview of visual guidance coordination on road-facing and driver-facing cameras.
In this task, as you can see from the figure, models are required to generate appropriate guidance coordination for each user based on the situation estimated from road-facing and driver-facing cameras by following the given instructions. We explain the details for each setting in the following subsections.

\subsubsection{Input}

In our task, models can receive three inputs: visual, audio, and text-based information. Thus, models need to jointly handle information from these three modalities. We introduce the characteristics and details of the information from each modality.

\paragraph{Visual Information}

Visual information plays a main role in understanding driving situations. 
In our task, models extract the visual information from road-facing and driver-facing cameras as video frames. The unique point of our task is the use of driver-facing cameras.
Unlike conventional research, our task requires models to understand the interaction between two types of cameras. Essentially, the number of cameras should not be restricted to a specific number. However, our work is the first attempt to handle both types of cameras, and the performance of conventional models in handling them is uncertain. Hence, in this work, we use only one road-facing and one driver-facing camera for simplification. 

\paragraph{Audio Information}

Audio information supports the insufficient visual information for models to understand driving situations. Moreover, it's also important to detect drivers' emotions. Both purposes are important for generating explanations for coaching. Thus, we provide internal car sounds gathered by microphones for this task.

\paragraph{Instruction}

Our task requires generating coaching explanations by following given text-based instructions. These instructions can change model behaviors. As we explained, the final goal of our task is to generate appropriate guidance coordination for each user. However, just generating explanations lacks evidence for the decision. Therefore, in addition to the explanation generation, we ask questions that are answerable with nouns as evidence via instructions to judge the appropriateness of model decisions. Table \ref{tab:instructions} shows both types of instructions. The event recognition type instructions require estimating predefined labels, whereas the open question type instructions require explanations.

\subsubsection{Output}

In this task, the output is text-only, and its content and format are decided by given instructions.  Table \ref{tab:instructions} also shows the required answers for each instruction type. As in the table, the responses are classified as labels or generated as explanations.

\subsection{Evaluation Metrics}
\label{subsec:metric}

To evaluate these two different types of instructions, we utilize the following evaluation metrics:

\paragraph{Event Recognition} Since event recognition is a kind of classification, we can use the common accuracy rate (AR) for the evaluation. To assess the accuracy of event recognition based on model responses, each item is queried in a chat-style format. Responses are categorized as True events when the models accurately match observed conditions and False events when they do not. The accuracy rate (AR) is calculated using the formula as follows:
\[
\text{AR} = \frac{\text{True Events}}{\text{False Events} + \text{True Events}}.
\]

\paragraph{Open Question} Different from event recognition, we cannot rely on AR in evaluating responses to open question-type instructions because of its explanation-styled output. Thus, instead of using AR, we incorporate reference-based automatic evaluation metrics, BLEU \cite{papineni-etal-2002-bleu} and BERTScore \cite{Zhang*2020BERTScore:}\footnote{\url{https://github.com/Tiiiger/bert_score}}, used in natural language generation tasks into our task. We can judge how models generate appropriate keywords in their explanations by using BLEU scores. Regarding the appropriateness of context in generated explanations, which sometimes differs from keyword matching, we can rely on BERTScores. Note that to avoid the fluctuation caused by tokenization, we use sacreBLEU \cite{post-2018-call}\footnote{\url{https://github.com/mjpost/sacrebleu}} in the evaluation.

\begin{table}[t]
    \centering
    \small
    \begin{tabular}{cccc}
        \toprule
        \multirow{2}{*}{\textbf{Split}} & \multirow{2}{*}{\textbf{Num. of Videos}} & \multicolumn{2}{c}{\textbf{Num. of Inst.}} \\
        \cmidrule{3-4}
        & & \textbf{ER} & \textbf{OQ} \\
        \midrule
        Train & 95 & --- & 190 \\
        Valid & 24 & --- & 48 \\
        Test  & 100 & 2,000 & 200 \\
        \bottomrule
    \end{tabular}
    \caption{Dataset statistics. Inst., ER, and OQ denote instructions, event recognition, and open questions, respectively.}
    \label{tab:statistics}
\end{table}

\section{Dataset Construction}
\label{sec:dataset}

Our dataset comprises synchronized video recordings from both road-facing and driver-facing RGB cameras. These recordings cover a diverse range of driving scenarios, including city roads and highways, various weather conditions, and different times of day. We included significant harsh driving events such as following too closely, harsh braking, and sharp turns, as well as major distractions, including phone usage, smoking, and signs of fatigue.

Preprocessing steps included frame extraction, resizing, and merging the frames from the road-facing and driver-facing cameras side by side to create a comprehensive view of each driving scenario.

After the preprocessing, we split the dataset into training, validation, and test sets. Table \ref{tab:statistics} shows the statistics for each split. In these sets, training and validation sets are used for training our models explained in \S\ref{sec:model_build} and the test set is used for evaluating this model.

\begin{figure}[t]
    \centering
    \includegraphics[width=\columnwidth]{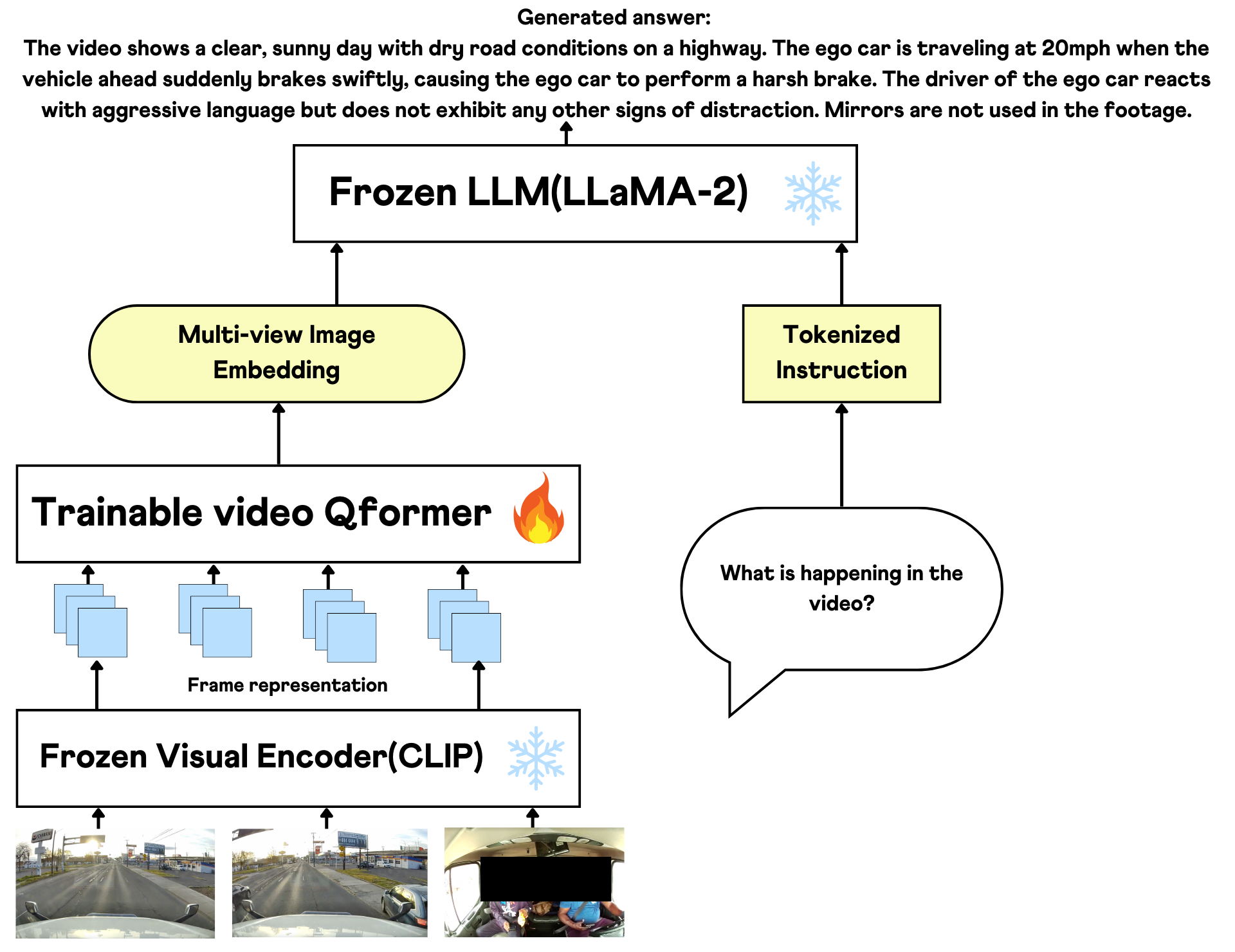}
    \caption{Integration of road-facing and driver-facing RGB camera footage.}
    \label{fig:integration}
\end{figure}

\section{Our Model: Multi-Frame Vision-Language Model}
\label{sec:model_build}

\subsection{Architecture}

Our Multi-Frame Vision-Language Model is built upon the Video-LLaMA \cite{zhang-etal-2023-video} framework, which integrates both visual and auditory content to enhance video understanding. In handling visual content, the model is required to receive multiple images simultaneously and integrate footage from both road-facing and driver-facing RGB cameras. Figure \ref{fig:integration} shows the integration overview. In addition to the video information, the audio information can support the model's decision. For this function, the model consists of two core components: the Vision-Language (VL) Branch and the Audio-Language (AL) Branch.

\paragraph{Vision-Language Branch}
The VL Branch utilizes a pre-trained visual encoder based on BLIP-2 \cite{10.5555/3618408.3619222}, specifically a ViT-G/14 model from EVA-CLIP \cite{fang2023eva}, to extract features from video frames. This encoder is frozen to retain the pre-trained knowledge. The extracted frame embeddings are then processed by a two-layer Video Q-Former and a frame embedding layer, which capture temporal relationships and transform the video embedding vectors into video query vectors. These vectors are of the same dimension as the text embeddings of the large language model (LLM) and are concatenated to the text embeddings as a video soft prompt, guiding the frozen LLM to generate text based on video content.

\paragraph{Audio-Language Branch}
To handle auditory content, the AL Branch employs a pre-trained audio encoder (ImageBind) to compute features from short audio segments. These segments are converted into spectrograms and mapped into dense vectors by the audio encoder. Similar to the Video Q-Former, the Audio Q-Former adds temporal information to the audio segments and generates fixed-length audio features. These features are then transformed into the embedding space of the LLM via a linear layer.

\subsection{Training}
\label{subsec:training}

We fine-tuned the Video-LLaMA model, \texttt{Video-LLaMA-2-13B-Finetuned}\footnote{\url{https://huggingface.co/DAMO-NLP-SG/Video-LLaMA-2-13B-Finetuned}}, using our created dataset (\S\ref{sec:dataset}), which comprises synchronized video recordings from both road-facing and driver-facing RGB cameras on 8 of NVIDIA A100 (80GB) with inhereting the original hyper-parameters.
This additional fine-tuning step ensures that the model is specifically adapted to our unique data, capturing the nuances of driving behavior and distractions and improving its performance in generating relevant and accurate coaching instructions. This three-stage approach ensures that the model effectively learns both general visual semantics, specific instructional tasks, and the particularities of our dataset.

During the fine-tuning, we froze the weights of the visual encoder and the large language model (LLM). The fine-tuning focused on updating the pre-trained weights of the video Qformer. This approach allows us to inject the necessary information to understand driving situations efficiently, reducing the GPU workload and making the process cost-effective. By freezing the weights, we leverage the pre-trained knowledge of the encoder and LLM, ensuring that only the specific layers related to video analysis are adjusted, which speeds up the training process and improves resource utilization.

\subsection{Inference}
\begin{figure}[t]
 \centering
    \includegraphics[width=0.9\linewidth]{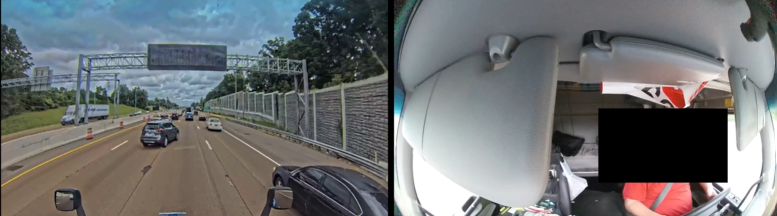}
    \caption{A frame of the inputted video to the models.}
    \label{fig:example video}
\end{figure}

During inference, the model processes multiple images simultaneously for cross-video analysis, integrating footage from both road-facing and driver-facing RGB cameras. This comprehensive understanding of driving behavior is facilitated by aligning the outputs from the vision-language model with a coaching database, where an LLM generates detailed descriptions and instructions for both drivers and managers. Figure \ref{fig:example video} illustrates our integration of footage from both camera types.

\begin{figure}[t]
    \centering
    \includegraphics[width=0.7\columnwidth]{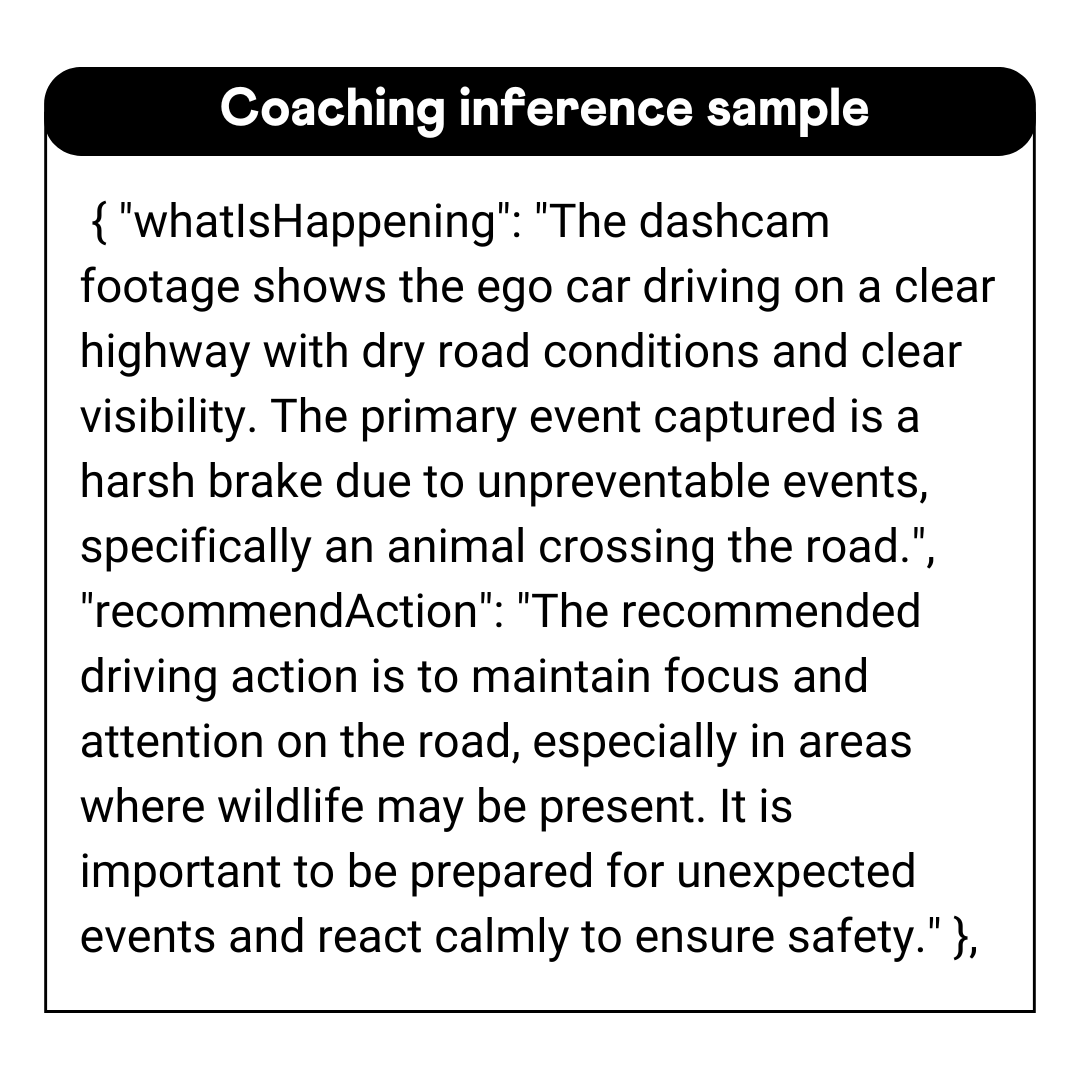}
    \caption{Coaching framework aligning vision-language model outputs with coaching instructions.}
    \label{fig:coaching_framework}
\end{figure}

Our coaching framework aligns the outputs from the vision-language model with the instructions in our coaching database. The framework utilizes an LLM to generate detailed situation descriptions and actionable instructions for both drivers and managers. Figure \ref{fig:coaching_framework} illustrates this process.

The framework ensures that the generated instructions are contextually relevant and detailed, facilitating effective coaching and enhancing the understanding of driving behaviors.

\section{Experiments}

\subsection{Settings}

\paragraph{Datasets}
We prepared videos that concatenate both driver-facing and road-facing video footages like in Figure \ref{fig:example video}, similar to our training dataset used for fine-tuning and our test set explained in \S\ref{sec:dataset}.

\paragraph{Models}
We compared the model with the original weight of \texttt{Video-LLaMA-2-13B-Finetuned} and the fine-tuned weight by our dataset (\S\ref{subsec:training}).

\paragraph{Metrics} As explained in \S\ref{subsec:metric}, we used AR for evaluating the performance of event recognition instructions and BLEU and BERTScore for evaluating that of open question instructions. Note that LLMs, not instruction-tuned on target tasks, sometimes generate malformed output due to their assistant-styled behavior. To deal with this problem, we manually removed unnecessary generation-like greetings that are not related to the required answers.

\begin{table}[t]
    \centering
    \small
    \begin{tabular}{cc}
        \toprule
        \textbf{Model} &  \textbf{AR} \\
        \midrule
        Video-LLaMA & 43.4 \\
        Ours & \textbf{67.7} \\
        \bottomrule
    \end{tabular}
    \caption{The results on event recognition instructions. The bold font indicates the best score.}
    \label{tab:results_event_recognition}
\end{table}

\begin{table}[t]
    \centering
    \small
    \begin{tabular}{ccccc}
   \toprule
   \multirow{2.5}{*}{\textbf{Model}} & \multirow{2.5}{*}{\textbf{BLEU}} & \multicolumn{3}{c}{\textbf{BERTScore}} \\
   \cmidrule{3-5}
         &      & \textbf{Precision} & \textbf{Recall} & \textbf{F1} \\
   \midrule
   Video-LLaMA & 0.9   &  0.836    & 0.840   & 0.837 \\
   Ours      &  \textbf{8.1}     &  \textbf{0.914}   & \textbf{0.885}   & \textbf{0.899} \\
   \bottomrule
    \end{tabular}
    \caption{The results on open question instructions. Notations are the same as in Table \ref{tab:results_event_recognition}.}
    \label{tab:results_open_question}
\end{table}

\subsection{Results}

Table \ref{tab:results_event_recognition} shows the results of models in event recognition instructions. These results indicate that without instruction-tuning, Video-LLaMA cannot generate the required specific keywords like our instruction-tuned model. 

Table \ref{tab:results_open_question} shows the results of models in open-question instructions. These results show the same tendency as the results of recognition instructions. Therefore, we can understand the importance of creating the dataset for instruction-tuning to improve the performance in this task.

\subsection{Analysis}

\begin{figure}[t]
 \centering
    \includegraphics[width=\columnwidth]{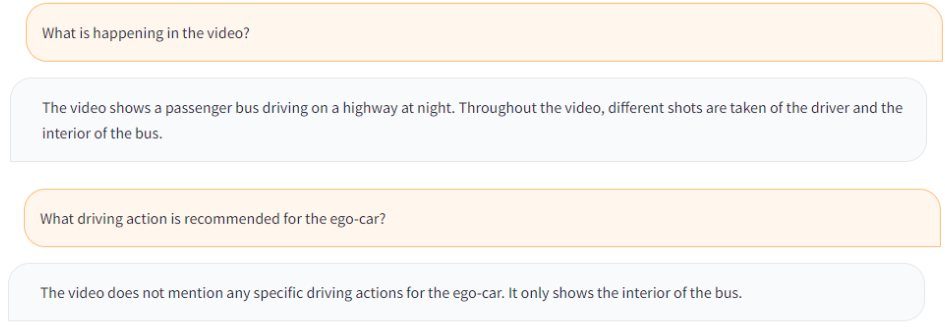}
    \caption{The example output from \texttt{Video-LLaMA-2-} \texttt{13B-Finetuned}.}
    \label{fig:example_video_1}
\end{figure}

\begin{figure}[t]
 \centering
    \includegraphics[width=\columnwidth]{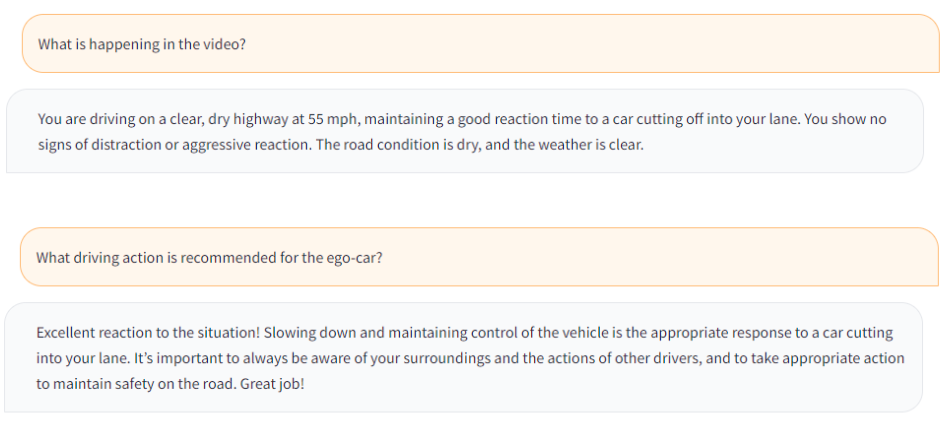}
    \caption{The example output from our fine-tuned model.}
    \label{fig:example_video_2}
\end{figure}

Figures \ref{fig:example_video_1} and \ref{fig:example_video_2} illustrate example responses to an open question regarding a specific video generated by two models trained in this study: the original \texttt{Video-LLaMA-2-13B-Finetuned} model and our additionally fine-tuned model.

By conducting the analysis of merged road-facing and driver-facing video footage, we identified several notable advantages over traditional visual language models in the context of vehicle dashcams. Specifically, the conventional model, \texttt{Video-LLaMA-2-13B-Finetuned}, frequently produced confusing and ambiguous analysis results when the merged video was used for inference. In contrast, our model not only shows the improved ability to capture and interpret events from the road-facing camera but also supplements this with driver-facing camera footage, effectively capturing driver reactions and responses to external events.

The ability to accurately recognize and respond to complex driving situations, such as lane cut offs, is another strength of our model. This temporal analysis capability allows our model to generate contextually relevant advice aimed at promoting safe driving behavior. By evaluating not only some, but a whole series of frames, our model is able to detect patterns and identify dynamic driving scenarios from the vehicle's trajectory.

Furthermore, our model was able to infer more accurate information about driving-related events observed in the video footage than \texttt{Video-LLaMA-2-13B-Finetuned}. This integrated approach provides more reliable and actionable insights by providing a holistic view of driving and allowing detailed, nuanced analysis of the driving environment without missing subtle cues, complex interactions, and causal relationships.

\section{Related Work}

\paragraph{Large-scale Vision Language Models (LVLMs)} such as BLIP-2 \cite{10.5555/3618408.3619222}, LLaVA \cite{liu2023visual} , mPLUG-Owl \cite{ye2024mplugowlmodularizationempowerslarge}, and Qwen-VL \cite{Qwen-VL}, etc., consist of large language models (LLMs) and a visual transformer (ViT) \cite{dosovitskiy2021an}-based pre-trained vision-encoder like CLIP \cite{radford2021learning} to handle various kinds of vision and language (V\&L) tasks including explanation generation like image review generation \cite{saito2024evaluatingimagereviewability} and artwork explanation \cite{hayashi2024artworkexplanationlargescalevision}. In the style of Video-LLaMA \cite{zhang-etal-2023-video}, we can expand LVLMs to cover video input by considering videos as multiple images. \citet{kamigaito-etal-2023-table} point out the discrepancy of knowledge between the language model and vision encoder caused by their separate training. To deal with this problem, we conduct instruction-tuning on our Multi-Frame Vision-Language Model.

\paragraph{Visual Instruction-tuning} \cite{liu2023visual} improves the performances of LVLMs by enhancing the ability to handle various tasks, whereas it requires instruction-tuning data. In a simple way, we can convert conventional datasets into instruction-tuning data by manually creating templates. In another way, \citet{liu2023visual} create their instruction-tuning data through automatic generation by GPT-4 \cite{openai2024gpt4technicalreport}. To maintain the quality of instruction-tuning, we manually created our instruction-tuning dataset.

\paragraph{Large Language Models for Autonomous Driving} are investigated by various approaches like \citet{xu2023drivegpt4,yang2023lidar,jin2023adapt,hu2023gaia1generativeworldmodel} as surveyed by \citet{zhou2023driving}. Different from their work, our proposed task targets coaching and requires the use of two different cameras, road-facing and driver-facing RGB cameras. Therefore, our work targets novel and uncovered fields that are challenging to be solved by conventional research direction.

\section{Conclusions}

This study presents a methodology for a visual language model in the context of risky behavior coaching and dashcam recording footage.

We've seen the emerging expansion of dashcam adaption for commercial fleets. Aleady 4 million fleets adapted technology and grew fast. 
They have event detection techniques but only for simple object detection, like phone usage or no seatbelt. In more complex road traffic-related events like near collisions, lane cut-offs, and harsh braking situations, there is a lack of efficient analysis methods, and managers need to review recordings all day to get insight and proceed with the coaching process. This problem is connected to delays in corrective action and fatal accidents. This experiment should change this situation with new video intelligence technology and driver coaching.

\section{Future Work}
There are multiple opportunities to enhance our models further.

One approach is to broaden the scope of our datasets. Currently, our model is trained using video and audio files recorded with dashcams installed on trucks. Given the increasing use of technologies such as LiDAR and radar in autonomous driving, incorporating these data types could significantly improve our model's accuracy and robustness. By integrating LiDAR and radar data, we can achieve a more comprehensive analysis of driving environments and behaviors, which would be particularly beneficial for detecting and understanding complex driving scenarios.

Another approach is to reduce the GPU workload through model optimization techniques such as model pruning, quantization, and efficient layer design. These techniques not only lower the overall cost of the computational environment but also enable real-time analysis, making our system more responsive and scalable. This improvement is crucial for deploying the model in real-world applications where timely feedback is essential for driver coaching and safety monitoring.

\section{Limitation}

Due to the limitation of the current multimodal large language models like Video-LLaMA \cite{zhang-etal-2023-video}, our instruction-tuned model, which is based on Video-LLaMA, cannot handle hours of video footage at once and may require splitting videos in practical situations.

\section{Ethical Consideration}

Since the drivers approved the use of their driving data for our research, there is no ethical consideration about our created dataset.

\bibliography{citations}

\end{document}